\documentclass[sigconf, review=false]{acmart}

\usepackage{booktabs} %
\usepackage{algorithm}
\usepackage{algorithmicx}
\usepackage{algpseudocode}
\usepackage{multirow}
\usepackage{multicol}
\usepackage{tabularx}
\usepackage{array}
\usepackage{xcolor}

\newcommand{\methodname}{PrinGS}
\newcommand{\datasetname}{PrinGS-Humans}

\newcommand{\stepA}{\protect\raisebox{-0.5ex}{\includegraphics[height=2.5ex]{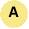}}}
\newcommand{\stepB}{\protect\raisebox{-0.5ex}{\includegraphics[height=2.5ex]{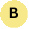}}}
\newcommand{\stepC}{\protect\raisebox{-0.5ex}{\includegraphics[height=2.5ex]{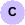}}}
\newcommand{\stepD}{\protect\raisebox{-0.5ex}{\includegraphics[height=2.5ex]{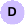}}}

\definecolor{projectpageblue}{HTML}{1484E9} 

\setcopyright{none}

\renewcommand\footnotetextcopyrightpermission[1]{}

\acmConference[ICVGIP2026]{17th Indian Conference on Computer Vision, Graphics and Image Processing}{December 21-25, 2026}{Kolkata, India}
\acmYear{2026}
\copyrightyear{2026}
\acmPrice{15.00}

\begin{document}
\title{\methodname{}: Privacy Preservation in Gaussian Splatting}

\author{Gaurav Behera}
\authornote{Equal Contribution.}
\affiliation{%
  \institution{IIIT Hyderabad, India}
  \country{}
}
\email{gaurav.behera@research.iiit.ac.in}

\author{Abhinav Raundhal}
\authornotemark[1]
\affiliation{%
  \institution{IIIT Hyderabad, India}
  \country{}
}
\email{abhinav.raundhal@students.iiit.ac.in}

\author{Archisha Panda}
\authornotemark[1]
\affiliation{%
  \institution{IIIT Hyderabad, India}
  \country{}
}
\email{archisha.panda@research.iiit.ac.in}

\author{Saumya Balina}
\authornotemark[1]
\affiliation{%
  \institution{IIIT Hyderabad, India}
  \country{}
}
\email{saumya.balina@students.iiit.ac.in}

\author{Gargi Shroff}
\authornotemark[1]
\affiliation{%
  \institution{IIIT Hyderabad, India}
  \country{}
}
\email{gargi.shroff@research.iiit.ac.in}

\author{Ponnurangam Kumaraguru}
\affiliation{%
  \institution{IIIT Hyderabad, India}
  \country{}
}
\email{pk.guru@iiit.ac.in}

\author{\texttt{\normalsize \color{projectpageblue} \href{https://privacypreserving3dgs.github.io}{PrivacyPreserving3DGS.github.io}}}

\renewcommand{\shortauthors}{}

\begin{abstract}
3D Gaussian Splatting (3DGS) has rapidly become the preferred representation for high-fidelity novel view synthesis and 3D scene reconstruction. However, the learned representation encodes sensitive visual information including human faces, license plates, documents, and computer screens, raising significant privacy concerns when trained splats are shared or deployed. In this work, we address the underexplored problem of privacy preservation in Gaussian Splatting by proposing a representation-level anonymization framework that directly anonymizes a trained splat. Given a trained splat and a user-provided reference view identifying sensitive content, our method localizes the corresponding regions in the underlying 3D representation and directly anonymizes them, ensuring that the sensitive information is absent from all subsequently rendered novel views while preserving the overall scene geometry and visual fidelity. We achieve this post-training anonymization in less than $1$ second. Furthermore, we extend our framework with a privacy-by-design training strategy that prevents sensitive content from being embedded into the representation during optimization, enabling the generation of privacy-preserving splats directly from training images. Our method achieves a remarkably low Face Similarity Measure (FSM) of $9.79\%$ at a retained Face Detection Rate (FDR) of $39.47\%$ and makes sensitive text completely unreadable with an OCR Character Error Rate of $1.00$. Our work aims to serve as the primary stepping stone towards privacy-aware 3D representations and motivates further research along these directions.
\end{abstract}



\begin{teaserfigure}
    \vspace{10pt}
    \centering
    \includegraphics[width=\textwidth]{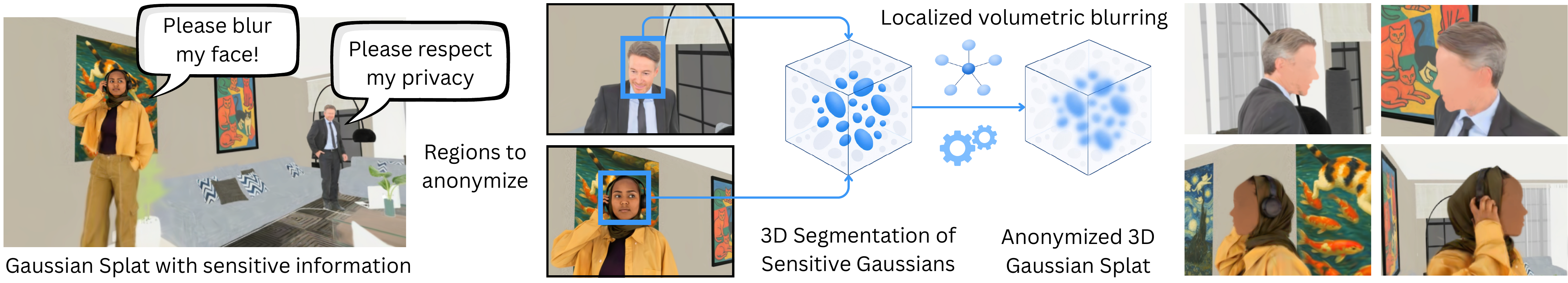}
    \caption{
    Privacy Preservation in Gaussian Splatting. Given only a trained splat and user-selected reference views, our method localizes the Gaussians corresponding to sensitive content and directly anonymizes the splat without requiring the original training images. The resulting privacy-preserving splat consistently hides faces across arbitrary novel viewpoints while preserving the fidelity of the remaining scene.
    }
    \Description{Overview of the proposed method.}
    \label{fig:teaser}
    \vspace{16pt}
\end{teaserfigure}

\maketitle

\vspace{3mm}

\section{Introduction}
\label{sec:intro}

\vspace{1.5mm}

The widespread adoption of 3D Gaussian Splatting (3DGS)~\cite{kerbl3Dgaussians} has enabled creating and sharing highly realistic 3D scene representations across applications such as robotics, autonomous navigation, digital twins, cultural heritage preservation, and virtual reality~\cite{zhu20243d, shimomura2025digital, jamil2025immersive}. 
While these representations excel at preserving visual fidelity, they also make it possible to recover information that users may expect to remain private, creating new challenges for the safe sharing and deployment of learned scene representations.
Since this information is embedded within the learned 3D representation it remains accessible from arbitrary novel viewpoints, introducing significant privacy risks and raising challenges for compliance with regulations such as the General Data Protection Regulation (GDPR)~\cite{EU_GDPR_2016} and the Digital Personal Data Protection (DPDP) Act~\cite{DPDPA2023}.

\newpage

An increasing number of splats are being publicly shared and deployed across real-world applications. A challenge arises when an individual notices their identity captured in a scene without consent and requests its removal to comply with privacy regulations. Retraining from scratch is computationally infeasible. Additionally, in many scenarios, the original training images used for reconstruction are unavailable, inaccessible, or prohibitively expensive to reacquire. Consequently, there is a pressing need for techniques that can anonymize existing splats without requiring access to the original training data.

\vspace{2mm}

A straightforward solution is to anonymize rendered images after novel-view synthesis. However, such image-space anonymization is inherently view-dependent and leaves the underlying 3D representation unchanged, allowing sensitive information to remain visible when the scene is rendered from different viewpoints. This limitation motivates a representation-level approach that directly anonymizes the learned splat, ensuring that sensitive information is removed across all novel views. We present \methodname{}, a framework for post-training privacy preservation in Gaussian Splatting. Given a trained splat and user-selected reference views containing sensitive content, our method localizes the corresponding Gaussians in the underlying 3D representation and directly anonymizes them. By operating on the representation itself rather than rendered images, sensitive information is removed from the splat, preventing its recovery from arbitrary novel viewpoints while preserving scene geometry and fidelity as shown in Figure~\ref{fig:teaser}.

\vspace{2mm}

While post-training anonymization is essential for existing splats, the rapid adoption of 3DGS is expected to result in an ever-growing number of new 3D scene representations. This presents an opportunity to incorporate privacy preservation directly into the reconstruction pipeline rather than addressing it retrospectively. Motivated by this, we further introduce a privacy-by-design framework that integrates our representation-level anonymization mechanism into the optimization process, preventing sensitive information from being encoded into the splat during training.

\vspace{2mm}

Existing 3DGS benchmarks predominantly lack privacy-sensitive content like faces, creating a critical gap that hinders the evaluation of privacy aware 3D scene reconstructions. To address this, we construct, \datasetname{}, a dataset of synthetic indoor scenes containing $9$ human subjects across varying demographics like skin tones and genders, establishing the first benchmark for systematic evaluation of privacy preservation of faces in Gaussian Splatting.

\vspace{2mm}

Furthermore, conventional rendering metrics primarily assess visual fidelity and do not capture privacy preservation. We therefore quantitatively evaluate our privacy-by-design approach using novel task-specific metrics based on Face Similarity Measure (FSM) and Face Detection Rate (FDR) for faces and OCR Character Error Rate to quantify the removal of textual sensitive information. In terms of facial privacy, our method achieves a Face Similarity Measure (FSM) of only $9.79\%$, representing a substantial reduction ($74.08\%$) compared to standard 3DGS ($83.87\%$). Furthermore, for textual content, our approach attains a Character Error Rate (CER) of $1.00$, indicating complete unreadability of private text, significantly outperforming baseline 3DGS, which yields a CER of $0.208$.

Our main contributions are summarized as follows:
\begin{itemize}
\item \textbf{\methodname{}}. We present the first framework for post-training privacy preservation in Gaussian Splatting that directly anonymizes the learned representation without requiring the original training images, preventing sensitive information from being recovered from novel viewpoints in just $1$ second.

\item \textbf{Privacy-by-Design}. We introduce a privacy-by-design extension that incorporates anonymization into the Gaussian Splatting optimization process, preventing sensitive information from being encoded during training.

\item \textbf{\datasetname{}}. To address the lack of static scenes with humans due to privacy reasons, we construct a benchmark of synthetic indoor scenes containing $9$ human subjects across varying demographics, enabling systematic evaluation of privacy-preserving Gaussian Splatting methods.

\item \textbf{Privacy Metrics.} We introduce novel evaluation protocols tailored to the 3D Gaussian Splatting domain, specifically utilizing the Face Similarity Measure (FSM), Face Detection Rate (FDR) and OCR Character Error Rate (CER) to rigorously quantify privacy preservation, demonstrating a $74.08\%$ reduction in FSM and a perfect CER of $1.00$.

\end{itemize}

\section{Related Work}
\label{sec:related_work}

\subsection{3D Gaussian Splatting}

Recent advancements in novel view synthesis have shifted from continuous volumetric fields (NeRF)~\cite{mildenhall2020nerf} to explicit and efficient representations. 3D Gaussian Splatting (3DGS)~\cite{kerbl3Dgaussians} models scenes as collections of anisotropic 3D Gaussians parameterized by spatial positions, covariance matrices, opacity, and spherical harmonic coefficients. This formulation bypasses costly volumetric ray marching, enabling real-time rendering and interactive scene exploration. 

Building upon the foundational 3DGS framework, subsequent research has branched into multiple directions to enhance rendering quality and efficiency. Notable quality improvements include anti-aliased formulations such as Mip-Splatting~\cite{yu2024mip} and specialized regularizations like AbsGS~\cite{ye2024absgs} that handle unbounded scenes and reduce artifacts. Another direction modifies the densification strategy such as 3DGS-MCMC~\cite{kheradmand20243d}.
To bridge the gap toward resource-constrained settings, methods like Taming 3DGS~\cite{mallick2024taming} introduce controlled densification to curb primitive redundancy. Building upon these efficiency measures, compact variants like Mini-Splatting~\cite{fang2024mini, fang2024mini2} introduce memory-efficient pruning and compression mechanisms. Despite these rapid developments in performance and visual fidelity, existing literature predominantly focuses on optimization efficiency and rendering quality while largely overlooking the security and privacy implications of capturing real-world environments.

A parallel line of work adds semantics to 3DGS by learning auxiliary per-Gaussian attributes during training, such as the SAM-distilled affinity features of SAGA~\cite{cen2025segment}, the identity encodings of Gaussian Grouping~\cite{ye2024gaussian}, and the feature fields or semantic tracing of Feature-3DGS~\cite{zhou2024feature} and GaussianEditor~\cite{chen2024gaussianeditor}. Since these attributes must be learned during optimization, such methods cannot be applied to an already-trained splat that does not carry them. \methodname{} instead localizes sensitive Gaussians directly from the rasterizer's transmittance-weighted contributions, and anonymizes them.

\subsection{Privacy Preservation in 2D Visual Data}

The rapid growth of visual data as images and videos on the internet has made privacy preservation an increasingly important research problem.
To mitigate the exposure of personally identifiable information (PII), regulatory frameworks have accelerated research into automated anonymization; however, existing efforts remain predominantly confined to the 2D domain.

For images, anonymization is achieved as a two-step process: detection or segmentation followed by blurring. For detection and segmentation, frameworks such as YOLO~\cite{hurtik2022poly, wang2024yolov1}, Faster R-CNN~\cite{ren2016faster}, Mask R-CNN~\cite{he2017mask}, and SAM~\cite{kirillov2023segment} are utilized to accurately localize sensitive regions. Following detection, privacy is enforced by applying conventional filtering and obfuscation mechanisms such as localized spatial blurring, pixelation, masking, or identity scrambling to protect sensitive regions~\cite{ribaric2016identification}.
In videos, privacy preservation builds upon these image-based techniques by incorporating temporal consistency through multi-object tracking algorithms~\cite{ciaparrone2020deep, senior2005enabling} to prevent identity leakage across successive frames.

\subsection{Privacy Preservation in 3D Scenes}
\label{baselines}

While privacy preservation has been extensively studied in 2D images and videos, extending these mechanisms to 3D scene representations remains a largely underexplored non-trivial task. In 3D environments, privacy preservation requires ensuring that sensitive content is completely anonymized across all arbitrary novel viewpoints rather than just a fixed set of frames.

Completely removing sensitive objects from a 3D scene is often impractical, as it degrades the semantic richness required by downstream applications. For example, autonomous driving systems require pedestrians to learn safe navigation, but their facial identities are unnecessary. Similarly, many robotics and digital twin applications benefit from preserving human presence while suppressing personally identifiable information. This motivates representation-level anonymization that retains scene semantics while removing identity-bearing visual attributes.

A naive baseline is to apply image-space anonymization directly to the training images prior to 3D reconstruction. However, this approach introduces fundamental limitations. First, masking or blurring training images degrades local feature quality, severely hindering feature matching and camera pose estimation in Structure-from-Motion (SfM) pipelines like COLMAP~\cite{schoenberger2016sfm, schoenberger2016mvs}, which serve as a critical prerequisite for 3DGS. Second, per-view image-space modifications often yield inconsistent appearances across viewpoints, destabilizing the optimization landscape for 3DGS (refer supplement for detailed analysis). 

To address these limitations, our work addresses privacy directly at the representation level. We are the first to formalize a privacy-preserving framework explicitly designed for 3DGS enabling view-consistent anonymization across arbitrary viewpoints while preserving scene geometry and rendering quality.

\begin{algorithm}[t]
\caption{3D Segmentation of Sensitive Gaussians}
\label{alg:segmentation}

\noindent\parbox{\linewidth}{%
  \textbf{Input:} Set of views $\mathcal{V}$, corresponding binary masks $\mathcal{M} = \{M_v\}_{v \in \mathcal{V}}$, 3D Gaussians $\mathcal{G} = \{g_1, \dots, g_N\}$, transmittance threshold $\tau_T$, global visibility threshold $\tau_v$ \\
  \textbf{Output:} Sensitive Gaussian subset $\mathcal{S}$
}
\begin{algorithmic}[1]
\State Initialize global visibility counters $C_i \leftarrow 0$ for all $g_i \in \mathcal{G}$
\For{each view $v \in \mathcal{V}$}
    \State \textit{\stepA \text{ }Single-view identification}
    \State Initialize $c_{\text{true}, i} \leftarrow 0$ and $c_{\text{false}, i} \leftarrow 0$ for all $g_i \in \mathcal{G}$
    \For{each pixel $p$ in view $v$}
        \For{each overlapping Gaussian $g_i$ at pixel $p$}
            \State $T_i \leftarrow \text{EvaluateTransmittance}()$
            \If{$T_i > \tau_T$}
                \If{$M_v(p) == 1$}
                    \State $c_{\text{true}, i} \leftarrow c_{\text{true}, i} + 1$
                \Else
                    \State $c_{\text{false}, i} \leftarrow c_{\text{false}, i} + 1$
                \EndIf
            \EndIf
        \EndFor
    \EndFor
    \State $\mathcal{S}_v \leftarrow \{ g_i \in \mathcal{G} \mid c_{\text{true}, i} > c_{\text{false}, i} \}$
    \State $\mathcal{S}'_v \leftarrow \text{IsolationForest}(\mathcal{S}_v)$ on $\mu_i$
    
    \State \textit{\stepB \text{ }Multi-view aggregation}
    \For{each $g_i \in \mathcal{S}'_v$}
        \State $C_i \leftarrow C_i + 1$
    \EndFor
\EndFor
\State \textbf{return} $\mathcal{S} \leftarrow \{ g_i \in \mathcal{G} \mid C_i \geq \tau_v \}$
\end{algorithmic}
\end{algorithm}

\section{Methodology}
\label{sec:method}

\begin{figure*}[t]
    \centering
    \includegraphics[width=\textwidth]{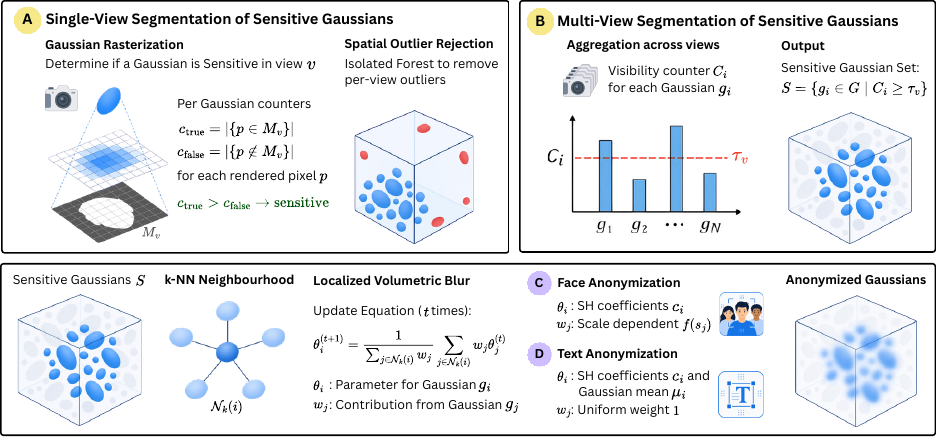}
    \caption{Overview of core components for our privacy preservation pipeline. \stepA \text{ }Single-view segmentation identifies sensitive Gaussians via rasterization against 2D masks combined with Isolation Forest outlier rejection. \stepB \text{ }Multi-view aggregation uses a visibility counter threshold to isolate the final set of sensitive Gaussians. Once identified, localized volumetric blur is applied using $k$-NN neighborhood. \stepC \text{ }Face anonymization modifies the spherical harmonic coefficients with scale-dependent weights. \stepD \text{ }Text anonymization, on the other hand, modifies both SH coefficients and spatial means with uniform weight. The final output is a splat with sensitive information anonymized.}
    \label{fig:method_setup}
\end{figure*}

\begin{figure}[t]
    \centering
    \includegraphics[width=\columnwidth]{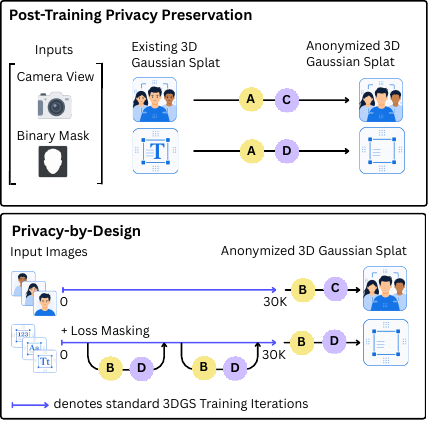}
    \caption{
    Overview of our two operational settings. Top: Post-Training Privacy Preservation takes an already-trained 3D Gaussian Splat along with a reference camera view and binary mask, to produce an anonymized splat without accessing original training images. Bottom: Privacy-by-Design integrates our anonymization pipeline directly into the training iterations along with loss masking to yield a anonymized splat.
    }
    \label{fig:method_pipeline}
\end{figure}

This section details our framework for privacy-preserving 3D Gaussian Splatting. We first introduce our approach for accurately segmenting sensitive Gaussians in 3D space (Section \ref{sec:segmentation}). Next, we define the localized anonymization operators for blurring sensitive content (Section \ref{sec:anonymization_formulation}). Finally, we describe how these techniques are applied in two distinct scenarios: modifying an existing Gaussian Splat post-training (Section \ref{sec:post_training}), and an end-to-end privacy-by-design pipeline that prevents sensitive features from being learned during optimization (Section \ref{sec:privacy_by_design}).
An overview of the proposed framework is illustrated in Figures~\ref{fig:method_setup} and~\ref{fig:method_pipeline}. Figure~\ref{fig:method_setup} presents the segmentation of sensitive Gaussians in 3D along with the localized anonymization operators applied to them. Figure~\ref{fig:method_pipeline} provides an overview of our two techniques: post-training anonymization of an existing splat and privacy-by-design training that prevents sensitive content from being encoded during optimization.

\subsection{3D Segmentation of Sensitive Gaussians}
\label{sec:segmentation}

To remove sensitive information from a 3D scene, we must first accurately segment in 3D the specific Gaussians responsible for rendering those regions. While heuristic COLMAP based 2D-to-3D projection methods \cite{STRinGS_2026_WACV} exist for 3D text segmentation, they often struggle with complex volumetric occlusions and background artifacts. To overcome these limitations, we integrate the segmentation directly into the differentiable rasterization pipeline, constructing a precise representation-level mapping by tracking splat contributions during alpha compositing stage of the rasterizer.

Our downstream tasks demand either single-view segmentation for post-training anonymization, or a robust accumulation throughout the entire scene to handle multiple views for privacy-by-design. To support these two scenarios, our segmentation process offers two corresponding methods: single-view identification \stepA\text{ }and multi-view aggregation \stepB, respectively. The overall logic is shown in Algorithm \ref{alg:segmentation}.

\stepA \text{ }For a given view $v$, let $M_v \in \{0, 1\}^{H \times W}$ be the binary mask indicating the sensitive 2D region. During the tile-based forward pass, for each pixel $p$ and each overlapping Gaussian $g_i$, we evaluate its accumulated transmittance $T_i$. If $T_i > \tau_T$, we atomically update one of two contribution counters for $g_i$: $c_{\text{true}}$ if $p \in M_v$ (i.e., $M_v(p) = 1$), and $c_{\text{false}}$ if $p \notin M_v$. A Gaussian $g_i$ is classified as sensitive for view $v$ if $c_{\text{true}} > c_{\text{false}}$. Additionally, as rendering is projective, the background geometries may coincidentally project into $M_v$. To robustly filter these spurious floaters, we apply an Isolation Forest algorithm~\cite{liu2008isolation} to the 3D spatial coordinates $\mu_i$ of all flagged Gaussians in the current frame, discarding spatial outliers.

\stepB \text{ }To aggregate these filtered identifications across multiple views, we maintain a global visibility counter $C_i$ for each Gaussian $g_i$, incrementing it whenever the Gaussian is marked sensitive in a view. The final sensitive subset $\mathcal{S}$ is defined as $\mathcal{S} = \{ g_i \in \mathcal{G} \mid C_i \geq \tau_{v} \}$, where $\tau_{v}$ is a global visibility threshold.

\subsection{Gaussian Anonymization Formulation}
\label{sec:anonymization_formulation}

Let the 3D Gaussian representation be denoted as $\mathcal{G} = \{g_1, g_2, \dots, g_N\}$, where each Gaussian $g_i$ is defined by its explicit parameters, including its spatial mean $\mu_i \in \mathbb{R}^3$, spherical harmonics (SH) coefficients $c_i$, and scaling vector $s_i$. Once the segmentation process identifies the sensitive subset $\mathcal{S} \subset \mathcal{G}$, we anonymize these targeted Gaussians by applying a localized volumetric blur to their parameters.

For each sensitive Gaussian $g_i \in \mathcal{S}$, we find its $k$-nearest spatial neighbors $\mathcal{N}_k(i)$. We then iteratively update a target parameter $\theta_i$ over $t$ steps using a weighted average of this neighborhood:
$$ \theta_i^{(t+1)} = \frac{1}{\sum_{j \in \mathcal{N}_k(i)} w_j} \sum_{j \in \mathcal{N}_k(i)} w_j \theta_j^{(t)} $$
The specific parameter $\theta_i$ updated and the weight $w_j$ applied depend on the type of sensitive content being anonymized.

\stepC \text{ }For face anonymization, preserving the fundamental 3D structure of the head is necessary to maintain scene coherence. Facial identity characteristics are predominantly tied to high-frequency visual details such as eyes, nose, and lips, which are naturally represented by Gaussians with smaller scales. Therefore, we restrict the anonymization operator strictly to the appearance parameters (the SH coefficients, $\theta_i = c_i$) and use a scale-dependent weight $w_j = f(s_j)$. This approach ensures that larger, structural background splats appropriately guide the color blending by stripping away identifying details while leaving the spatial geometry intact.

\stepD \text{ }For text anonymization, readable characters are mostly planar, and preserving their local geometric structure can inadvertently leave readable physical cues. We therefore expand the anonymization to target both appearance and geometry ($\theta_i = \{c_i, \mu_i\}$) using a uniform, unweighted filter ($w_j = 1$). This effectively dissolves both the text's coloring and its physical shape, collapsing distinct character geometries into unreadable text.

\subsection{Post-Training Privacy Preservation}
\label{sec:post_training}

Our framework can seamlessly modify an already optimized 3D Gaussian Splat without requiring the original training data. The inputs for this post-training setting are simply the existing 3D Gaussian Splat, a single reference camera view capturing the sensitive region and its corresponding 2D binary privacy mask.

Because this setting relies on a single user-provided view, we utilize the single-view segmentation process \stepA \text{ }to identify the sensitive Gaussians. Once isolated, the corresponding anonymization operator is applied directly to the static representation: facial regions undergo scale-weighted appearance blurring \stepC, while text regions receive uniform appearance and structural blurring \stepD. By altering the core representation, the sensitive information is permanently erased across all novel views.

\subsection{Privacy-by-Design}
\label{sec:privacy_by_design}

While post-training edits are effective, retrospectively modifying a highly optimized scene may sometimes leave back traces of private information, if sensitive Gaussians are deeply entangled with the scene structure. Our Privacy-by-Design approach mitigates this by integrating anonymization directly into the standard 3DGS training pipeline, ensuring sensitive features are never encoded from the input images.

\textbf{Face Anonymization:} For facial regions, interrupting the optimization gradients mid-training degrades the underlying geometric fidelity of the head, which is needed for view-dependent coherence. Therefore, we apply anonymization only after the standard training iterations (e.g., 30K) when the geometry of the faces has converged.
This involves multi-view segmentation \stepB \text{ }followed by applying the scale-weighted facial anonymization operator \stepC \text{ }to the final splat. This preserves the properly learned geometry while completely removing high-frequency facial details.

\textbf{Text Anonymization:} Text geometries are highly sensitive to even weak gradients, meaning sharp, readable artifacts can easily form during training. To prevent this, we explicitly exclude text regions from the photometric loss computation using masked non-sensitive background regions:
$$ \mathcal{L} = (1 - \lambda)\mathcal{L}_1 (I \odot \bar{M_v}, \hat{I} \odot \bar{M_v}) + \lambda \mathcal{L}_{\text{D-SSIM}}(I \odot \bar{M_v}, \hat{I} \odot \bar{M_v}) $$
where $I$ and $\hat{I}$ denote the ground-truth and rendered images, respectively, and $\bar{M_v} = 1 - M_v$ represents the inverted mask for view $v$. Furthermore, to suppress any accidental readable structures formed by unsupervised floating geometry, we periodically interrupt the training loop to apply the multi-view text segmentation \stepB \text{ }followed by the uniform structural anonymization operator \stepD. This continuous intervention ensures readable text is never optimized.

\section{Results and Experiments}
\label{sec:results}

We evaluate our approach in both the post-training privacy preservation and privacy-by-design settings. The former demonstrates the ability to anonymize existing Gaussian Splats without requiring access to the original training images, while the latter enables quantitative evaluation through the availability of ground-truth images. We compare against standard 3D Gaussian Splatting and analyze both qualitative and quantitative aspects of privacy preservation.

\subsection{Experimental Setup}

All experiments are conducted on a single NVIDIA RTX 3090 Ti GPU with 24\,GB VRAM. Following the 3DGS~\cite{kerbl3Dgaussians} literature, every eighth frame is reserved for evaluation while the remaining frames are used for optimization. For all experiments, we use the optimization strategies for faces and text as described in Section~\ref{sec:method}. For detailed implementation details, refer to the supplement.

\begin{figure*}[t]
    \centering
    {\normalsize
    \begin{tabularx}{\linewidth}{*{5}{>{\centering\arraybackslash}X}}
        Reference View &
        Novel View 1 &
        Novel View 2 &
        Novel View 3 &
        Novel View 4
    \end{tabularx}
    }
    \includegraphics[width=\textwidth]{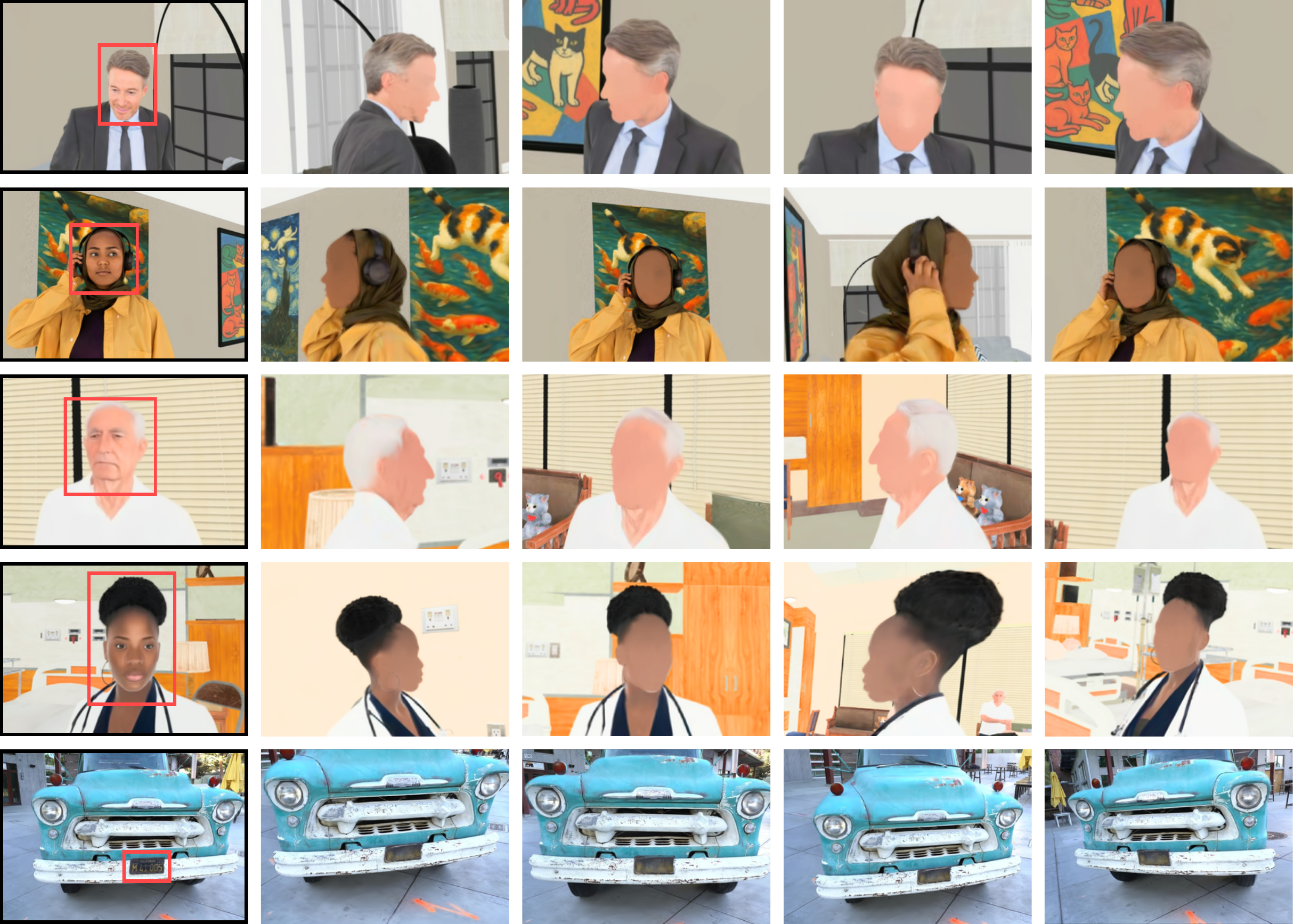}
    \caption{
    \textbf{Qualitative results for post-training anonymization on scenes from our \datasetname{} dataset (rows 1-4) and Tanks and Temples~\cite{Knapitsch2017} (row 5).}
    Starting from a single user-selected \emph{reference view} (left), our method directly anonymizes the underlying Gaussian representation without requiring access to the original training images. The remaining columns show novel views rendered after representation-level anonymization. Sensitive facial and textual information is consistently removed across arbitrary viewpoints, demonstrating that privacy preservation naturally propagates through the edited 3D representation rather than being limited to the selected view. 
    Notice the geometry of the faces being preserved across all novel views, with the silhouette sharply defining face contours. (Best seen on screen). 
    }
    \label{fig:post_training_results}
\end{figure*}

\subsection{Datasets}

\subsubsection{\datasetname{} Dataset.}
Existing Gaussian Splatting benchmarks contain no human subjects suitable for evaluating facial privacy preservation. We therefore construct a synthetic benchmark in Blender comprising two indoor scenes containing $9$ unique human subjects~\cite{renderpeople2025, restore50_sketchfab}. The dataset includes subjects with diverse appearances, including varying skin tones, ages, genders, hairstyles, facial hair making the benchmark representative of real-world humans.

\subsubsection{Text Privacy Dataset.}
For evaluating textual privacy, we select five scenes containing readable text from existing public benchmarks. Specifically, we use the \textit{Truck}, \textit{Barn}, and \textit{M60} scenes from Tanks and Temples~\cite{Knapitsch2017} together with two scenes from the DL3DV-10K benchmark~\cite{ling2024dl3dv} which we call \textit{Car} and \textit{Piano}. These scenes contain diverse private text instances like number plates, house numbers, and brand names with varying font sizes, colours, orientations, and viewing distances, providing a challenging benchmark for text anonymization.

\subsection{Evaluation Protocol}

\subsubsection{Post-training Anonymization}

The post-training setting assumes that only a trained Gaussian Splat is available, without access to the original calibrated training images. Given an arbitrary reference view rendered using the SIBR viewer~\cite{sibr2020}, the user specifies the sensitive region to be anonymized, which is segmented in 2D using SAM3~\cite{carion2025sam}. Our method then localizes the corresponding Gaussians in the learned representation and performs representation-level anonymization as described in Section~\ref{sec:post_training}. Since the modification is performed directly in the Gaussian representation, the anonymization naturally propagates to all rendered viewpoints.

We demonstrate the effectiveness of our post-training framework through extensive qualitative results across multiple scenes and viewpoints in Figure~\ref{fig:post_training_results}, with a controlled quantitative evaluation in the supplement.

\subsubsection{Privacy-by-Design}

We next consider the privacy-by-design setting where sensitive regions are anonymized during optimization. Given input images, sensitive regions including faces, license plates, documents, and other privacy-sensitive content, are segmented using SAM3~\cite{carion2025sam} with user-provided text prompts. These masks are used by the optimization pipeline to output a privacy-preserving Gaussian representation, as described in Section~\ref{sec:privacy_by_design}. Since ground-truth images are available in this setting, both qualitative and quantitative evaluation become possible.

\begin{figure}[t]
    \centering
    {\normalsize
    \begin{tabularx}{\linewidth}{*{3}{>{\centering\arraybackslash}X}}
        Ground Truth &
        3DGS~\cite{kerbl3Dgaussians} &
        \textbf{\methodname{} (Ours)} \\
    \end{tabularx}
    }
    \includegraphics[width=\columnwidth]{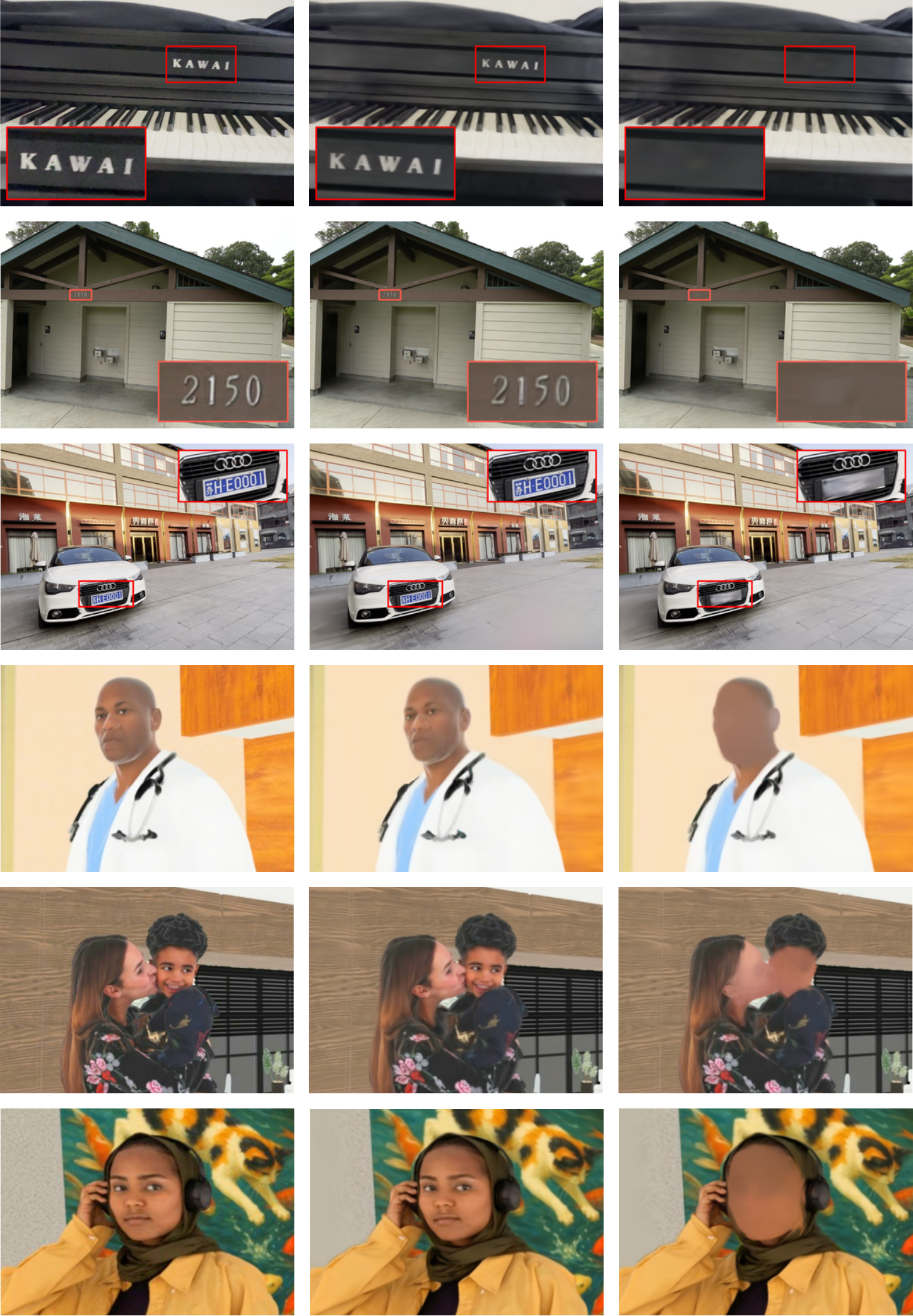}
    \caption{
    \textbf{Qualitative comparison of privacy-by-design training on scenes from DL3DV-10K~\cite{ling2024dl3dv} (rows 1,3), Tanks and Temples~\cite{Knapitsch2017} (row 2), and our \datasetname{} Dataset (rows 4-6).}
    From left to right: ground truth image, novel rendered images from 3DGS, and our method \methodname{}. 
    While 3DGS reconstructs sensitive content, our approach prevents such information from being encoded during training, effectively anonymizing the scene while preserving the appearance of non-sensitive regions.
}
    \label{fig:privacy_by_design_results}
\end{figure}

\subsection{Evaluation Metrics}

Conventional rendering metrics such as PSNR, SSIM, and LPIPS~\cite{zhang2018unreasonable} primarily measure rendering fidelity and are not designed to quantify privacy preservation. To address this, we employ three task-specific metrics: Face Similarity Measure (FSM), Face Detection Rate (FDR), and OCR Character Error Rate (CER).

\subsubsection{Face Similarity Measure (FSM)}
To evaluate the anonymization of facial recognition features, we use the industry-standard InsightFace framework on the novel-view rendered images.

The evaluation pipeline utilizes an SCRFD-10GF~\cite{guo2021sample} detector to scan the entire rendered image and detect faces. Following detection, we extract 512-dimensional embeddings via ArcFace~\cite{deng2018arcface} (ResNet-50) CNN backbone. FSM is then quantified using Cosine Similarity between the embeddings of the ground-truth and anonymized faces. A lower cosine similarity indicates successful anonymization.

\subsubsection{Face Detection Rate (FDR)}
Since a ground-truth face with no corresponding detection in the render is assigned a similarity of $0$, FSM alone rewards a method that destroys the face entirely rather than anonymizing it. We therefore also report the Face Detection Rate (FDR), the fraction of ground-truth faces across the evaluation views that remain detectable by the SCRFD-10GF~\cite{guo2021sample} detector in the anonymized render.

\subsubsection{OCR Character Error Rate (CER)}
To evaluate text readability, we utilize the Google Cloud Vision API~\cite{google_cloud_vision} as a robust Optical Character Recognition (OCR) tool. Following STRinGS~\cite{STRinGS_2026_WACV}, we quantify text readability using recall-based Character Error Rate (CER), computed via the Levenshtein edit distance between text extracted from the ground-truth images and the anonymized renders.
In the context of privacy preservation, a higher CER demonstrates that the underlying textual geometry has been successfully destroyed.

\subsection{Privacy-by-Design Results}

Figure~\ref{fig:privacy_by_design_results} qualitatively compares the rendered images produced by standard 3D Gaussian Splatting and our privacy-preserving framework against the ground truth. Our approach effectively removes sensitive facial and textual information while maintaining high rendering quality for the remaining scene content.

\begin{table}[t]
\centering
\caption{
\textbf{Face Similarity Measure and Face Detection Rate on our \datasetname{} benchmark.}
Lower FSM indicates stronger anonymization, while higher FDR indicates that facial geometry is preserved. 3DGS reconstructs facial identity, whereas our method substantially suppresses identity-related features while keeping the faces detectable.
}
\label{tab:fdr_fsm}

\normalsize
\setlength{\tabcolsep}{8pt}

\begin{tabular}{lcccc}
\toprule
& \multicolumn{2}{c}{3DGS} & \multicolumn{2}{c}{\textbf{Ours}} \\
\cmidrule(lr){2-3}\cmidrule(lr){4-5}
Scene & FDR $\uparrow$ & FSM $\downarrow$ & FDR $\uparrow$ & FSM $\downarrow$ \\
\midrule
House    & 91.67 & 78.68 & 36.11 & 6.55  \\
Hospital & 97.50 & 88.55 & 42.50 & 12.70 \\
\midrule
Overall  & 94.74 & 83.87 & \textbf{39.47} & \textbf{9.79} \\
\bottomrule
\end{tabular}

\end{table}

Table~\ref{tab:fdr_fsm} reports the Face Similarity Measure (FSM) and Face Detection Rate (FDR) on \datasetname{}. Because subjects appear across a varying number of evaluation frames, per-subject FSM and FDR values are computed and aggregated via weighted averages to derive scene-level metrics. Standard 3DGS yields high overall similarity at an FDR of $94.74\%$, indicating severe privacy leakage, as the faces are both present and identifiable. In contrast, our method drastically reduces embedding similarity to an overall average of $9.79\%$ while still retaining an FDR of $39.47\%$, meaning the anonymized faces are detected as faces, yet the recognizer can no longer match them to the correct identity. 
FDR is based on a face detector which takes both facial features and geometry into account for inference, justifying the $39.47\%$ FDR of retained geometry and lost facial features.
Quantitative metrics comparing \methodname{} with baselines mentioned in Subsection \ref{baselines} are presented in the supplement. Our method is the only one that drops FSM while maintaining a reasonable FDR.

\begin{table}[t]
\centering
\caption{
\textbf{OCR Character Error Rate (OCR-CER) comparison.}
Higher values indicate stronger anonymization, with a value of $1.0$ corresponding to complete failure of text recognition. Our method consistently suppresses readable text across all scenes, achieving the maximum OCR-CER, whereas standard 3DGS reconstructs text with substantially lower error.
}
\label{tab:ocr_results}

\normalsize
\setlength{\tabcolsep}{8pt}

\begin{tabular}{lccccc}
\toprule
OCR-CER $\uparrow$ & Truck & Barn & M60 & Piano & Car\\
\midrule
3DGS & 0.500& 0.028& 0.207& 0.149& 0.158\\
Ours & 1.00& 1.00& 1.00& 1.00& 1.00\\
\bottomrule
\end{tabular}

\vspace{-5pt}
\end{table}

Table~\ref{tab:ocr_results} reports the CER across all text scenes. Our approach substantially increases OCR error rates compared to standard Gaussian Splatting, indicating successful suppression of readable text. Specifically, while standard 3DGS yields an average CER of $0.208$ across five evaluation scenes, our method achieves a perfect recall-based CER of $1.00$, completely disrupting text recognition.

In summary, our quantitative and qualitative evaluations confirm that the proposed framework successfully masks sensitive facial and textual information across arbitrary novel viewpoints.

\subsection{Ablation Studies}

\subsubsection{Runtime Analysis}

\begin{table}[b]
\centering
\caption{
Runtime analysis of post-training anonymization. We report the total number of Gaussians in each scene (\#G), the number of Gaussians identified for anonymization (\#SensitiveG), Neighbourhood Size ($k$), Anonymization Iterations ($t$), and the runtime of our post-training anonymization stage. P1--P4 index the individual human subjects within a scene, following the per-subject numbering used in the supplement.
}
\label{tab:runtime}

\setlength{\tabcolsep}{6pt}
\normalsize

\begin{tabular}{lccccc}
\toprule
Scene & \#G & \#SensitiveG & $k$ & $t$ & Time (s)\\
\midrule
House (P1) & 592K & 1531& 25& 30& 0.70\\
House (P2) & 592K& 1213& 40& 30& 0.57\\
Hospital (P4) & 538K& 1334& 40& 30& 0.62\\
Hospital (P2) & 538K& 1127& 40& 30& 0.52\\
Truck & 2052K& 2551& 100& 1& 0.24\\

\bottomrule
\end{tabular}

\end{table}

Finally, we report the runtime of our post-training anonymization framework in Table~\ref{tab:runtime} along with Number of Sensitive Gaussians, Neighbourhood Size ($k$) and Anonymization Iterations ($t$). Since our method directly operates on the learned Gaussian representation without requiring retraining, anonymization can be performed efficiently on existing splats, in less than $1$ second, making the approach practical for real-world deployment.

\subsubsection{Neighborhood size ($k$) and Number of Anonymization Iterations ($t$)}
Our method updates each sensitive Gaussian using information aggregated from its neighboring Gaussians. We evaluate different neighborhood sizes $k$ and report the resulting Face Similarity Measure (FSM) while simultaneously studying the effect of the number of anonymization iterations $t$.

\begin{figure}[t]
    \centering
    \includegraphics[width=\columnwidth]{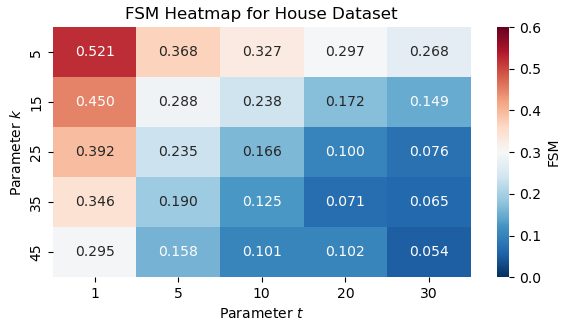}
    \caption{
    \textbf{Impact of neighborhood size ($k$) and anonymization iterations ($t$) on FSM.} Evaluation on the House dataset shows that higher values of $k$ and $t$ yield lower FSM scores (stronger privacy preservation), with gains eventually saturating.
}
\label{fig:ablation_heatmap}
\vspace{-10pt}
\end{figure}

As illustrated in the heatmap in Figure~\ref{fig:ablation_heatmap} for the House dataset, increasing both the neighborhood size $k$ and the optimization duration $t$ consistently yields better quantitative privacy results (lower FSM scores). However, after a certain threshold, these quantitative improvements gradually saturate. Beyond this convergence point, further increases in $k$ and $t$ transition from being strictly necessary for privacy gains to serving as a control for visual aesthetics, allowing users to strike the right balance between local color retention and a more uniform blending. 

\section{Conclusion}
\label{sec:conclusion}

We presented a unified framework for privacy preservation in 3D Gaussian Splatting through representation-level anonymization. Unlike existing approaches that rely on modifying input images prior to reconstruction, our method directly edits the learned Gaussian representation, enabling privacy preservation even when the original training images are unavailable. This post-training capability allows existing Gaussian Splats to be anonymized without requiring retraining, while the same formulation naturally extends to a privacy-by-design setting during Gaussian Splat optimization. To facilitate systematic evaluation, we introduced benchmarks for both facial and textual privacy and evaluated our approach using task-specific metrics based on Face Re-Identification similarity and OCR Character Error Rate, complemented by qualitative evaluation of post-training anonymization. Experimental results demonstrate that our method consistently removes sensitive information across novel viewpoints while preserving the visual fidelity of the remaining scene. As 3D scene representations become increasingly prevalent in applications such as robotics, digital twins, autonomous systems, and immersive visualization, incorporating privacy into their design will become increasingly important. We hope this work serves as a step toward practical privacy-aware 3D representations and motivates further research on privacy-preserving methods, benchmarks, and evaluation protocols for emerging 3D scene understanding and reconstruction systems.

\newpage

\bibliographystyle{unsrtnat}
\bibliography{references}

@Article{kerbl3Dgaussians,
      author       = {Kerbl, Bernhard and Kopanas, Georgios and Leimk{\"u}hler, Thomas and Drettakis, George},
      title        = {3D Gaussian Splatting for Real-Time Radiance Field Rendering},
      journal      = {ACM Transactions on Graphics},
      number       = {4},
      volume       = {42},
      month        = {July},
      year         = {2023},
      url          = {https://repo-sam.inria.fr/fungraph/3d-gaussian-splatting/}
}

@article{zhu20243d,
  title={3d gaussian splatting in robotics: A survey},
  author={Zhu, Siting and Wang, Guangming and Kong, Xin and Kong, Dezhi and Wang, Hesheng},
  journal={arXiv preprint arXiv:2410.12262},
  year={2024}
}

@article{shimomura2025digital,
  title={Digital-twin imaging based on descattering Gaussian splatting},
  author={Shimomura, Suguru and Yamanouchi, Kazuki and Tanida, Jun},
  journal={Optics Express},
  volume={33},
  number={14},
  pages={29351--29359},
  year={2025},
  publisher={Optica Publishing Group}
}

@article{jamil2025immersive,
  title={Immersive heritage through gaussian splatting: A new visual aesthetic for reality capture},
  author={Jamil, Osama and Brennan, AnnMarie},
  journal={Frontiers in Computer Science},
  volume={7},
  pages={1515609},
  year={2025},
  publisher={Frontiers}
}

@jurisdiction{EU_GDPR_2016,
  title        = {Regulation (EU) 2016/679 of the European Parliament and of the Council of 27 April 2016 on the protection of natural persons with regard to the processing of personal data and on the free movement of such data, and repealing Directive 95/46/EC (General Data Protection Regulation)},
  author       = {{European Parliament and Council of the European Union}},
  year         = {2016},
  journal      = {Official Journal of the European Union},
  volume       = {L 119},
  pages        = {1--88},
  url          = {http://europa.eu},
  urldate      = {2026-07-27}
}

@legislation{DPDPA2023,
  title     = {The Digital Personal Data Protection Act, 2023},
  number    = {Act No. 22 of 2023},
  author    = {{Parliament of India}},
  year      = {2023},
  date      = {2023-08-11},
  url       = {https://www.indiacode.nic.in/bitstream/123456789/22037/1/a2023-22.pdf},
  booktitle = {The Gazette of India}
}

@inproceedings{mildenhall2020nerf,
 title={NeRF: Representing Scenes as Neural Radiance Fields for View Synthesis},
 author={Ben Mildenhall and Pratul P. Srinivasan and Matthew Tancik and Jonathan T. Barron and Ravi Ramamoorthi and Ren Ng},
 year={2020},
 booktitle={ECCV},
}

@inproceedings{yu2024mip,
  title={Mip-splatting: Alias-free 3d gaussian splatting},
  author={Yu, Zehao and Chen, Anpei and Huang, Binbin and Sattler, Torsten and Geiger, Andreas},
  booktitle={Proceedings of the IEEE/CVF conference on computer vision and pattern recognition},
  pages={19447--19456},
  year={2024}
}

@inproceedings{ye2024absgs,
  title={Absgs: Recovering fine details in 3d gaussian splatting},
  author={Ye, Zongxin and Li, Wenyu and Liu, Sidun and Qiao, Peng and Dou, Yong},
  booktitle={Proceedings of the 32nd ACM international conference on multimedia},
  pages={1053--1061},
  year={2024}
}

@article{kheradmand20243d,
  title={3d gaussian splatting as markov chain monte carlo},
  author={Kheradmand, Shakiba and Rebain, Daniel and Sharma, Gopal and Sun, Weiwei and Tseng, Yang-Che and Isack, Hossam and Kar, Abhishek and Tagliasacchi, Andrea and Yi, Kwang Moo},
  journal={Advances in Neural Information Processing Systems},
  volume={37},
  pages={80965--80986},
  year={2024}
}

@inproceedings{mallick2024taming,
  title={Taming 3dgs: High-quality radiance fields with limited resources},
  author={Mallick, Saswat Subhajyoti and Goel, Rahul and Kerbl, Bernhard and Steinberger, Markus and Carrasco, Francisco Vicente and De La Torre, Fernando},
  booktitle={SIGGRAPH Asia 2024 Conference Papers},
  pages={1--11},
  year={2024}
}

@inproceedings{fang2024mini,
  title={Mini-splatting: Representing scenes with a constrained number of gaussians},
  author={Fang, Guangchi and Wang, Bing},
  booktitle={European conference on computer vision},
  pages={165--181},
  year={2024},
  organization={Springer}
}

@article{fang2024mini2,
  title={Mini-splatting2: Building 360 scenes within minutes via aggressive gaussian densification},
  author={Fang, Guangchi and Wang, Bing},
  journal={arXiv preprint arXiv:2411.12788},
  year={2024}
}

@article{hurtik2022poly,
  title={Poly-YOLO: higher speed, more precise detection and instance segmentation for YOLOv3},
  author={Hurtik, Petr and Molek, Vojtech and Hula, Jan and Vajgl, Marek and Vlasanek, Pavel and Nejezchleba, Tomas},
  journal={Neural Computing and Applications},
  volume={34},
  number={10},
  pages={8275--8290},
  year={2022},
  publisher={Springer}
}

@article{wang2024yolov1,
  title={YOLOv1 to YOLOv10: The fastest and most accurate real-time object detection systems},
  author={Wang, Chien-Yao and Liao, Hong-Yuan Mark},
  journal={arXiv preprint arXiv:2408.09332},
  year={2024}
}

@article{ren2016faster,
  title={Faster R-CNN: Towards real-time object detection with region proposal networks},
  author={Ren, Shaoqing and He, Kaiming and Girshick, Ross and Sun, Jian},
  journal={IEEE transactions on pattern analysis and machine intelligence},
  volume={39},
  number={6},
  pages={1137--1149},
  year={2016},
  publisher={IEEE}
}

@inproceedings{he2017mask,
  title={Mask r-cnn},
  author={He, Kaiming and Gkioxari, Georgia and Doll{\'a}r, Piotr and Girshick, Ross},
  booktitle={Proceedings of the IEEE international conference on computer vision},
  pages={2961--2969},
  year={2017}
}

@inproceedings{kirillov2023segment,
  title={Segment anything},
  author={Kirillov, Alexander and Mintun, Eric and Ravi, Nikhila and Mao, Hanzi and Rolland, Chloe and Gustafson, Laura and Xiao, Tete and Whitehead, Spencer and Berg, Alexander C and Lo, Wan-Yen and others},
  booktitle={Proceedings of the IEEE/CVF international conference on computer vision},
  pages={4015--4026},
  year={2023}
}

@article{ribaric2016identification,
  title={De-identification for privacy protection in multimedia content: A survey},
  author={Ribaric, Slobodan and Ariyaeeinia, Aladdin and Pavesic, Nikola},
  journal={Signal Processing: Image Communication},
  volume={47},
  pages={131--151},
  year={2016},
  publisher={Elsevier}
}

@article{ciaparrone2020deep,
  title={Deep learning in video multi-object tracking: A survey},
  author={Ciaparrone, Gioele and S{\'a}nchez, Francisco Luque and Tabik, Siham and Troiano, Luigi and Tagliaferri, Roberto and Herrera, Francisco},
  journal={Neurocomputing},
  volume={381},
  pages={61--88},
  year={2020},
  publisher={Elsevier}
}

@article{senior2005enabling,
  title={Enabling video privacy through computer vision},
  author={Senior, Andrew and Pankanti, Sharath and Hampapur, Arun and Brown, Lisa and Tian, Ying-Li and Ekin, Ahmet and Connell, Jonathan and Shu, Chiao Fe and Lu, Max},
  journal={IEEE Security \& Privacy},
  volume={3},
  number={3},
  pages={50--57},
  year={2005},
  publisher={IEEE}
}

@inproceedings{schoenberger2016sfm,
    author={Sch\"{o}nberger, Johannes Lutz and Frahm, Jan-Michael},
    title={Structure-from-Motion Revisited},
    booktitle={Conference on Computer Vision and Pattern Recognition (CVPR)},
    year={2016},
}

@inproceedings{schoenberger2016mvs,
    author={Sch\"{o}nberger, Johannes Lutz and Zheng, Enliang and Pollefeys, Marc and Frahm, Jan-Michael},
    title={Pixelwise View Selection for Unstructured Multi-View Stereo},
    booktitle={European Conference on Computer Vision (ECCV)},
    year={2016},
}

@InProceedings{STRinGS_2026_WACV,
  author    = {Raundhal, Abhinav and Behera, Gaurav and Narayanan, P. J. and Sarvadevabhatla, Ravi Kiran and Tapaswi, Makarand},
  title     = {STRinGS: Selective Text Refinement in Gaussian Splatting},
  booktitle = {Proceedings of the Winter Conference on Applications of Computer Vision (WACV)},
  month     = {March},
  year      = {2026},
}

@article{Knapitsch2017,
    author    = {Arno Knapitsch and Jaesik Park and Qian-Yi Zhou and Vladlen Koltun},
    title     = {Tanks and Temples: Benchmarking Large-Scale Scene Reconstruction},
    journal   = {ACM Transactions on Graphics},
    volume    = {36},
    number    = {4},
    year      = {2017},
}

@inproceedings{ling2024dl3dv,
  title={Dl3dv-10k: A large-scale scene dataset for deep learning-based 3d vision},
  author={Ling, Lu and Sheng, Yichen and Tu, Zhi and Zhao, Wentian and Xin, Cheng and Wan, Kun and Yu, Lantao and Guo, Qianyu and Yu, Zixun and Lu, Yawen and others},
  booktitle={Proceedings of the IEEE/CVF Conference on Computer Vision and Pattern Recognition},
  pages={22160--22169},
  year={2024}
}

@misc{sibr2020,
   author       = "Bonopera, Sebastien and Esnault, Jerome and Prakash, Siddhant and Rodriguez, Simon and Thonat, Theo and Benadel, Mehdi and Chaurasia, Gaurav and Philip, Julien and Drettakis, George",
   title        = "sibr: A System for Image Based Rendering",
   year         = "2020",
   url          = "https://gitlab.inria.fr/sibr/sibr_core"
}

@article{carion2025sam,
  title={Sam 3: Segment anything with concepts},
  author={Carion, Nicolas and Gustafson, Laura and Hu, Yuan-Ting and Debnath, Shoubhik and Hu, Ronghang and Suris, Didac and Ryali, Chaitanya and Alwala, Kalyan Vasudev and Khedr, Haitham and Huang, Andrew and others},
  journal={arXiv preprint arXiv:2511.16719},
  year={2025}
}

@inproceedings{zhang2018unreasonable,
  title={The unreasonable effectiveness of deep features as a perceptual metric},
  author={Zhang, Richard and Isola, Phillip and Efros, Alexei A and Shechtman, Eli and Wang, Oliver},
  booktitle={Proceedings of the IEEE conference on computer vision and pattern recognition},
  pages={586--595},
  year={2018}
}

@article{guo2021sample,
  title={Sample and computation redistribution for efficient face detection},
  author={Guo, Jia and Deng, Jiankang and Lattas, Alexandros and Zafeiriou, Stefanos},
  journal={arXiv preprint arXiv:2105.04714},
  year={2021}
}

@inproceedings{deng2018arcface,
title={ArcFace: Additive Angular Margin Loss for Deep Face Recognition},
author={Deng, Jiankang and Guo, Jia and Niannan, Xue and Zafeiriou, Stefanos},
booktitle={CVPR},
year={2019}
}

@misc{google_cloud_vision,
  author       = {{Google LLC}},
  title        = {Cloud Vision API (Optical Character Recognition)},
  howpublished = {\url{https://google.com}},
  year         = {2026},
  note         = {Accessed: July 27, 2026}
}

@misc{renderpeople2025,
  author       = {{Renderpeople GmbH}},
  title        = {{Renderpeople}: Scanned 3D People Models},
  howpublished = {\url{https://renderpeople.com/}},
  year         = {2025},
  note         = {Accessed: 2026-07-28}
}

@misc{restore50_sketchfab,
  author       = {{restore50}},
  title        = {3D Models Profile on Sketchfab},
  howpublished = {\url{https://sketchfab.com/restore50/models}},
  year         = {2026},
  note         = {Accessed: 2026-07-28}
}

@inproceedings{cen2025segment,
  title={Segment any 3d gaussians},
  author={Cen, Jiazhong and Fang, Jiemin and Yang, Chen and Xie, Lingxi and Zhang, Xiaopeng and Shen, Wei and Tian, Qi},
  booktitle={Proceedings of the AAAI conference on artificial intelligence},
  volume={39},
  number={2},
  pages={1971--1979},
  year={2025}
}

@inproceedings{ye2024gaussian,
  title={Gaussian grouping: Segment and edit anything in 3d scenes},
  author={Ye, Mingqiao and Danelljan, Martin and Yu, Fisher and Ke, Lei},
  booktitle={European conference on computer vision},
  pages={162--179},
  year={2024},
  organization={Springer}
}

@inproceedings{chen2024gaussianeditor,
  title={Gaussianeditor: Swift and controllable 3d editing with gaussian splatting},
  author={Chen, Yiwen and Chen, Zilong and Zhang, Chi and Wang, Feng and Yang, Xiaofeng and Wang, Yikai and Cai, Zhongang and Yang, Lei and Liu, Huaping and Lin, Guosheng},
  booktitle={2024 IEEE/CVF Conference on Computer Vision and Pattern Recognition (CVPR)},
  pages={21476--21485},
  year={2024},
  organization={IEEE}
}

@inproceedings{zhou2024feature,
  title={Feature 3dgs: Supercharging 3d gaussian splatting to enable distilled feature fields},
  author={Zhou, Shijie and Chang, Haoran and Jiang, Sicheng and Fan, Zhiwen and Zhu, Zehao and Xu, Dejia and Chari, Pradyumna and You, Suya and Wang, Zhangyang and Kadambi, Achuta},
  booktitle={2024 IEEE/CVF Conference on Computer Vision and Pattern Recognition (CVPR)},
  pages={21676--21685},
  year={2024},
  organization={IEEE}
}

@inproceedings{liu2008isolation,
  title={Isolation forest},
  author={Liu, Fei Tony and Ting, Kai Ming and Zhou, Zhi-Hua},
  booktitle={2008 eighth ieee international conference on data mining},
  pages={413--422},
  year={2008},
  organization={Ieee}
}

\end{document}